\documentclass{article}

\PassOptionsToPackage{numbers, compress}{natbib}

\usepackage[preprint]{neurips_2023}

\usepackage[utf8]{inputenc} 
\usepackage[T1]{fontenc}    
\usepackage{hyperref}       
\usepackage{url}            
\usepackage{booktabs}       
\usepackage{amsfonts}       
\usepackage{nicefrac}       
\usepackage{microtype}      
\usepackage{xcolor}         
\usepackage{algorithm}
\usepackage{algpseudocode}
\usepackage{graphicx}
\usepackage{amsmath}
\usepackage{listings}
\usepackage{xcolor}
\usepackage{mdframed}
\usepackage{amssymb}
\usepackage{subcaption}
\usepackage{listings}
\usepackage{xcolor}

\definecolor{backcolour}{rgb}{0.95, 0.95, 0.92}
\definecolor{codegray}{rgb}{0.5, 0.5, 0.5}
\definecolor{codeblue}{rgb}{0, 0, 1}
\definecolor{codegreen}{rgb}{0, 0.6, 0}
\lstdefinestyle{mystyle}{
    backgroundcolor=\color{backcolour},   
    commentstyle=\color{codegray},
    keywordstyle=\color{codeblue},
    numberstyle=\color{codeblue},
    stringstyle=\color{codegreen},
    basicstyle=\ttfamily\fontsize{10}{12}\selectfont,
    breakatwhitespace=false,         
    breaklines=true,                 
    captionpos=b,                    
    keepspaces=true,                                   
    showspaces=false,                
    showstringspaces=false,
    showtabs=true,                  
    tabsize=2,
}
\usepackage{array}
\newcommand{\LocalUpdate}{\textbf{InstructionTuning}}
\newcommand{\Aggregate}{\textbf{Aggregate}}

\newcommand{\shortname}{FedIT}

\usepackage{tabularx}
\usepackage{booktabs}
\usepackage{arydshln}

\title{Towards Building the Federated GPT: \\  Federated Instruction Tuning }

\author{
  Jianyi Zhang\textsuperscript{1}\thanks{Equal contribution, work done at Duke University.},
  Saeed Vahidian\textsuperscript{1}\footnotemark[1], Martin Kuo\textsuperscript{1}\footnotemark[1],
  Chunyuan Li\textsuperscript{2}, \\
  \textbf{Ruiyi Zhang}\textsuperscript{3},
  \textbf{Tong Yu}\textsuperscript{3},
  \textbf{Yufan Zhou}\textsuperscript{3},
  \textbf{Guoyin Wang}\textsuperscript{4}, \textbf{Yiran Chen}\textsuperscript{1}\thanks{The condensed version of this paper was accepted by ICASSP 2024. This work is supported in part by the grants NSF-2112562, NSF-2332744 and ARO W911NF-23-2-0224.} \\
  \textsuperscript{1} Duke University,
  \textsuperscript{2} Microsoft Research,
  \textsuperscript{3} Adobe Research,
  \textsuperscript{4} Amazon
}

\begin{document}

\maketitle




\begin{abstract}

While "instruction-tuned" generative large language models (LLMs) have demonstrated an impressive ability to generalize to new tasks, the training phases heavily rely on large amounts of diverse and high-quality instruction data (such as ChatGPT and GPT-4). Unfortunately, acquiring high-quality instructions, especially when it comes to human-written instructions, can pose significant challenges both in terms of cost and accessibility. Moreover, concerns related to privacy can further limit access to such data, making the process of obtaining it a complex and nuanced undertaking. Consequently, this hinders the generality of the tuned models and may restrict their effectiveness in certain contexts. To tackle this issue, our study introduces a new approach called \textbf{Fed}erated \textbf{I}nstruction \textbf{T}uning (FedIT), which leverages federated learning (FL) as the learning framework for the instruction tuning of LLMs. This marks the first exploration of FL-based instruction tuning for LLMs.  This is especially important since text data is predominantly generated by end users. For example, collecting extensive amounts of everyday user conversations in different languages can be a useful approach to improving the generalizability of LLMs, allowing them to generate authentic and natural responses. Therefore, it is imperative to design and adapt FL approaches to effectively leverage these  users' diverse instructions stored on local devices, while preserving privacy and ensuring data security. In the current paper, by conducting widely used GPT-4 auto-evaluation, we demonstrate that by exploiting the heterogeneous and diverse sets of instructions on the client's end with the proposed framework FedIT, we improved the performance of LLMs compared to centralized training with only limited local instructions. Further, in this paper, we developed a Github repository named {\textit{Shepherd}}. This repository offers a foundational framework for exploring federated fine-tuning of LLMs using heterogeneous instructions across diverse categories. The framework is designed for ease of use, adaptability, and scalability to accommodate large datasets. Additionally, it facilitates the seamless integration of novel algorithms and configurations, making it a convenient tool for researchers and practitioners in the NLP community.

\end{abstract}

\section{Introduction}

Large Language Models (LLMs) have become ubiquitous in natural language processing (NLP) \cite{brown2020language,devlin2018bert,  radford2018improving, radford2019language}, where one single model can perform well on various language tasks, including established tasks such as text generation, machine translation, and question answering, as well as novel application-oriented tasks in human daily life \cite{ding2019cognitive,sanh2021multitask,wang2018glue}. To align LLM to follow human intents, instruction-tuning has been proposed by fine-tuning LLM on instruction-following data \cite{ouyang2022training,wang2022self,wei2021finetuned}. 
Though instruction-tuning has demonstrated great effectiveness in improving the zero and few-shot generalization capabilities of LLM, its performance on real-world tasks is contingent on the \textit{quantity, diversity,} and \textit{quality} of the collected instructions \cite{mishra-etal-2022-cross,wang2022self}. The process of collecting these instructions can be expensive \cite{alpaca,wang2022self}.

Beyond the commonly acknowledged constraints of time and labor expenses, the increasing awareness of data sensitivity highlights a significant challenge in acquiring extensive and high-quality instructions \cite{balunovic2022lamp,gupta2022recovering,klymenko2022differential}. For instance, collecting vast amounts of daily conversations from users is a valuable means of providing guidance for LLMs, enabling them to generate authentic and genuine responses. However, privacy concerns may hinder users from sharing their conversations, resulting in a limited quantity of instructions that are not fully representative of the target population. Likewise, many companies treat their instructions as proprietary assets that are closely guarded. They are reluctant to share their instructions with external parties, as they often contain confidential and proprietary information that is critical to their success and profitability \cite{gurman2023samsung}. For example, pharmaceutical companies rely on meticulously-crafted instructions that may include details about the chemical composition of new drugs and the results of clinical trials \cite{kraljevic2004accelerating}. Hence, the sensitive nature of these instructions can pose significant challenges when utilizing traditional centralized approaches for instruction tuning.





We aim to tackle these challenges by exploring the potential of federated learning (FL) as a promising solution \cite{mcmahan2016communication}. This collaborative learning technique enables many clients to learn a shared model jointly without sharing their sensitive data. In particular, in our proposed federated instruction-tuning, clients initially download a global LLM from a central server and subsequently compute local model updates using their respective local instructions. These local updates are then transmitted back to the server, where they are aggregated and integrated to update the global LLM. Given that clients often have limited computational resources in comparison to traditional centralized training cloud servers, which can utilize thousands of GPUs to fully fine-tune all parameters of LLMs, we resort to parameter-efficient tuning  techniques. This leads to a significant decrease in the computational and communication demands, as it reduces the number of trainable parameters on each device.


%
Thus, our proposed framework enables efficient utilization of the computational resources available on local edge devices, which are commonly accessible, as well as their diverse local instructions. This eliminates the need for depending on large cloud servers to fine-tune LLMs.
Our major contributions are summarized as follows:






\begin{itemize}


\item We make the first attempt to leverage FL for instruction tuning (FedIT) of LLMs. In the predominant instruction tuning of LLMs, acquiring access to extensive and high-quality instructions can present significant obstacles due to the associated costs and privacy concerns. In this work, we show that we are able to circumvent the above-mentioned limitations by exploiting the diverse sets of available instructions (especially in cross-device FL where the number of clients is of the order of billion) from the users in the FL system. To make deploying LLM within an FL setup viable in terms of communication and computation cost, we suggest the clients exploit the parameter-efficient tuning (PETuning) methods.

\item A comprehensive study is conducted on the heterogeneity within the FL instruction tuning. We employ the GPT-4 auto-evaluation method, which has been widely utilized in related research \cite{vicuna2023, peng2023instruction}, to demonstrate the effectiveness of our FedIT approach in enhancing response quality by leveraging diverse available instructions. We discuss potential avenues for future research to improve the FL-based LLM fine-tuning methods for practical deployment. 
    

 \item We have developed and released a GitHub repository called {\textit{Shepherd}}\footnote{\url{https://github.com/JayZhang42/FederatedGPT-Shepherd}}, which has been designed to provide ease of customization and adaptability, thereby offering benefits for future research endeavors in this field.

\end{itemize}

\section{Related Work}

\subsection{Instruction tuning of Large Language Models}
Instruction tuning has emerged as a simple yet effective approach to enhance the generalizability of LLMs for complicated real-world tasks. This research area has recently gained increasing attention, particularly since the introduction of FLAN \cite{wei2021finetuned} that demonstrates significant zero-shot performance, and Instruct-GPT~\citep{ouyang2022training} that aligns GPT-3 \cite{brown2020language} to follow human intents via supervised tuning and RLHF~\cite{Christiano2017DeepRL,stiennon2020learning}. The development of Instruct-GPT has been instrumental in the success of ChatGPT \cite{openai-chatgpt} and GPT-4 \cite{GPT4report}.


In general, current research efforts can be broadly classified into two main categories based on the source of instructions: (1) human-annotated task prompts and feedback~\citep{ouyang2022training}, and (2) machine-generated instruction-following data. For the latter, self-instruct \cite{wang2022self} is utilized, where a strong teacher LLM is considered to generate a comprehensive collection of instructional data that a student LLM can then utilize to gain alignment capabilities. Thanks to the recently open-sourced LLM LLaMA~\cite{touvron2023llama}, which has demonstrated performance on par with proprietary LLMs such as GPT-3, the open-source community now has ample opportunities to actively explore promising solutions to build their own LLMs capable of following language and multimodal instructions~\cite{vicuna2023,liu2023llava,peng2023instruction, alpaca, xu2023baize,llamaadapter2023}. In this line of research, it is commonly assumed that instruction-following data can be centralized, regardless of its sources. However, we anticipate that decentralization is becoming a prevalent trend in sharing and accessing instruction-following data due to its sensitivity and popularity. As such, we propose the first attempt to address this issue using FL.





\paragraph{Parameter-Efficient Fine-Tuning (PEFT)}
The fine-tuning of LLMs aims to optimize LLMs while minimizing the computational and storage demands associated with the training process. Various innovative methods have been proposed to achieve this goal, each with distinctive characteristics, including LoRA \cite{hu2021lora},  P-Tuning \cite{liu2021gpt},  Prefix Tuning \cite{li-liang-2021-prefix,liu2021p}, Prompt Tuning \cite{lester-etal-2021-power}. We suggest interested readers to refer to the DeltaPaper repository \footnote{\url{https://github.com/thunlp/DeltaPapers}} and the Delta Tuning paper \cite{ding2022delta} for a comprehensive understanding of the advanced PEFT methods. We consider LoRA  in our FL framework due to its promising performance in recent studies on instruction tuning, including Alpaca-lora \footnote{\url{https://github.com/tloen/alpaca-lora}} and Baize \cite{xu2023baize}. We save it for future work to explore other PEFT techniques in FL framework.


\subsection{Federated Learning in NLP Tasks}

Federated Learning \cite{mcmahan2017communication} is a decentralized and collaborative machine learning technique that enables data to remain on user devices. Significant research efforts have focused on addressing privacy and heterogeneity challenges and developing advanced FL methods \cite{kairouz2021advances,  mahlool2022comprehensive,Vahidian-flis, nextgenFL,zhang2022next}. These advancements include designing optimization methods with improved aggregation performance (\cite{chen2020fedmax,du2022rethinking,hao2021towards,2020arXiv200300295R, sahu2018federated, Vahidian-ICDCS,zhu2021data}, increasing the framework's robustness against adversarial attacks \cite{Sun_2021_CVPR}, devising effective client selection mechanisms \cite{ Cho2020ClientSI,goetz2019active,vahidian-curr-FL,fedcbs}, enhancing personalization capabilities \cite{diao2020heterofl,li2020lotteryfl, vahidian-pacfl-2022,flop_qian2021}, and boosting the overall efficiency of FL systems \cite{oortFanLai,hermes,MeDNN,reisizadeh2020straggler,tang2022fade}.

Furthermore, recent research has explored the application of FL to NLP tasks, such as Language Modeling \cite{hard2018federated,yang2018applied}, Text Classification \cite{ chaudhary2022federated,lit2022federated}, Sequence Tagging \cite{ge2020fedner,jana-biemann-2021-investigation}, and Dialogue Generation \cite{lifedassistant,lu2021federated}. Several open benchmarks and repositories support the study of federated NLP tasks, including the Leaf benchmark \cite{caldas2018leaf}, FedNLP benchmark \cite{lin2021fednlp}, FedML \cite{he2020fedml}, FedScale \cite{ fedscale-icml22}, and FATE \cite{fate}. Recent research has also highlighted the importance of pretraining models for federated learning  \cite{chen2022pre,tan2022federated,FedBERT,weller2022pretrained}, as they offer a more powerful initialization for training instead of starting from scratch. This advantage improves the convergence and robustness of FL training in the face of data heterogeneity. 
Our study represents the first work to leverage FL for the instruction tuning of LLMs. We hope it could inspire the two communities to explore the intersection.

\section{Federated Instruction Tuning}

In light of our review in FL and instruction tuning, we now proceed to our Federated Instruction Tuning (FedIT). This section is structured as follows: first, we introduce the overall framework in subsection \ref{FLframework}. Subsequently, we delve into the intricacies of this framework and examine the heterogeneity of the instruction dataset in subsection \ref{heterogeneity}. Following this, we discuss the parameter-efficient technique, LoRA, that is integrated into our framework and explore its connections with other federated learning algorithms. Lastly, in subsection \ref{Shepherdframework}, we present our Python-based GitHub repository framework, ``\textit{Shepherd}" to facilitate research in federated instruction studies.

\subsection{An Overview of \shortname{}} \label{FLframework}

\begin{figure*}[t!]
    \centering
    \includegraphics[width=1.01\textwidth]{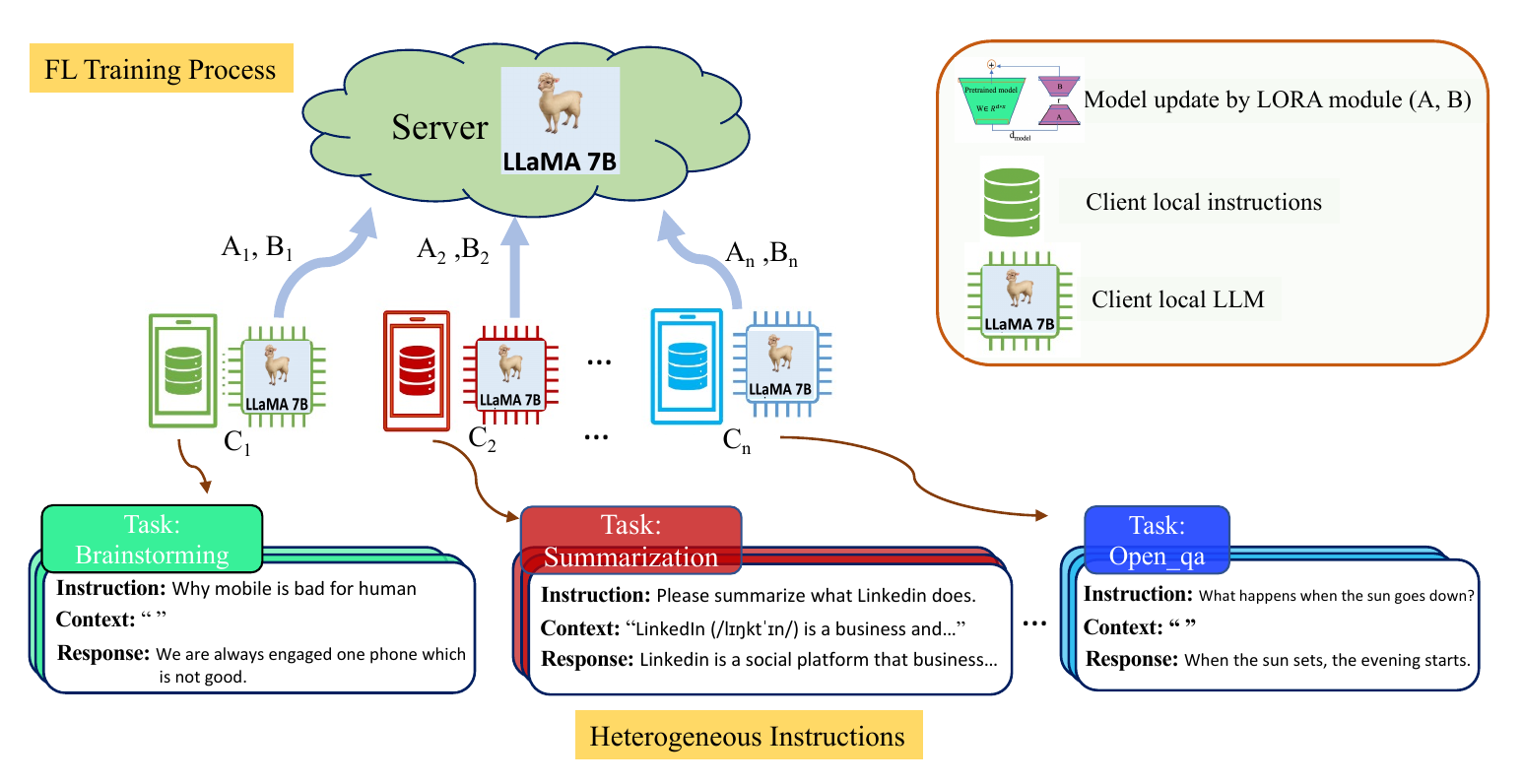}
    \caption{The framework of Federated Instruction Tuning (\shortname)}
    \label{fig:framework}
    \vspace{-5mm}
\end{figure*}

Drawing on the successful application of FL in various machine learning domains to offer privacy protection, we introduce the FedIT framework. By harnessing the advantages of FL and PEFT, our framework enables secure and cost-effective LLM instruction tuning. The overall framework, illustrated in Figure \ref{fig:framework} and Algorithm \ref{federated_learning_algorithm}, involves two primary components: local training operations on the client side and scheduling and aggregation operations on the server side, which work together to ensure efficient training.

Our FedIT framework for instruction tuning is designed to address the challenges of collecting high-quality data and ensuring data privacy by keeping the instructions on the local devices throughout the process. By ensuring data sensitivity protection, we can encourage more clients to participate in the federated instruction tuning. Consequently, the combined instruction dataset from all clients can encompass a broader range of topics, tasks, and valuable information, as clients may come from different areas and possess domain-specific expertise. This FL approach enables our framework to effectively adapt to diverse and evolving instruction datasets, resulting in more robust and generalized LLM performance.  Moreover, our FedIT methodology incorporates a parameter-efficient finetuning technique, known as LoRA, to facilitate local training. This method reduces computational and communication overheads for local edge devices that have limited system resources. As a result, we can leverage the computational capabilities of a multitude of distributed local edge devices that are often disregarded in conventional centralized instruction tuning. This feature enhances the scalability of our FedIT solution, enabling it to address large-scale instructional tuning challenges effectively.

The framework assigns an LLM to each client and performs client selection to determine which clients will participate in local instruction tuning. During instruction tuning, clients use their local instruction dataset to update a small, trainable adapter that is added to the pre-trained model weights. This approach reduces the cost of fine-tuning and makes it compatible with the limited computational resources of local devices. Upon completion, clients send the updated adapter back to the server, which aggregates the received adapters' parameters and conducts another round of client selection. This iterative process continues until convergence is achieved.

We also wish to emphasize the importance of client selection. In real-world settings, not all clients may be accessible for local instruction tuning, as local device processors may be occupied by other tasks. Hence, client selection can come into play to better simulate a real-world scenario. The server can actively choose clients for training based on their distinct instructions and computational resources, thus improving the overall efficiency of the FL framework by identifying clients that best represent the overall data distribution.


\begin{algorithm}
\caption{Federated Instruction Turning (FedIT)}\label{federated_learning_algorithm}
\small
\begin{algorithmic}
\item \hspace{-4mm}
\noindent \colorbox[rgb]{1, 0.95, 1}{
\begin{minipage}{0.9\columnwidth}

\textbf{\textbf{Initialization:}} each client's initial global large language model with parameters $\boldsymbol{w}$ and a lightweight adapter with parameters $\Delta \boldsymbol{w}^{(0)}$, client index subset $\mathcal{M}=\varnothing$, $K$ communication rounds, $k=0$,

\end{minipage}
}
\item \hspace{-4mm}
\colorbox[gray]{0.95}{
\begin{minipage}{0.9\columnwidth}
\item  \textbf{Training}

\item     \hspace*{\algorithmicindent} \textbf{while} $k \leq K$ \textbf{do}

\item     \hspace*{\algorithmicindent} \quad Server updates $\mathcal{M}$ using specific strategies 
            \Comment{ \textbf{\color{blue}  Select clients for local training}}

\item     \hspace*{\algorithmicindent} \quad \textbf{for} $n\in \mathcal{M}$ \textbf{in parallel do} \Comment{\textbf{\color{blue}Parameter-efficient finetuning on local instructions dataset} }

\item     \hspace*{\algorithmicindent} \quad \quad  Client freeze the  LLM and update the adapter weights with $\Delta \boldsymbol{w}^{(k)}$

\item     \hspace*{\algorithmicindent} \quad \quad  $\Delta \boldsymbol{w}_n^{(k+1)} \gets \LocalUpdate(\Delta \boldsymbol{w}_n^{(k)})$
\item     \hspace*{\algorithmicindent} \quad \textbf{end For}
\item     \hspace*{\algorithmicindent} \quad $\Delta \boldsymbol{w}^{(k+1)}\leftarrow \Aggregate(\Delta \boldsymbol{w}_n^{(k+1)})$ for $n \in\mathcal{M}$ \Comment{\textbf{\color{blue} Aggregate the adapters at Server}}
\item     \hspace*{\algorithmicindent} \quad  $k \gets k+1$
\item     \hspace*{\algorithmicindent}  \textbf{end while}
\end{minipage}
}
\item \hspace{-4mm}
\colorbox[rgb]{0.95, 0.98, 1}{
\begin{minipage}{0.9\columnwidth}

\item  \textbf{Outcome ($m, \theta^t_g$):}


\item     Derive the final adapter with parameters $\Delta \boldsymbol{w}^{(K)}$ and the global LLM with parameters $\boldsymbol{w}$ 
\end{minipage}
}
\end{algorithmic}
\end{algorithm}

\subsection{Heterogeneity of Instructional Data} \label{heterogeneity}

Beyond the practical benefits of FedIT, our research makes a unique contribution by presenting a scenario for instruction tuning of LLMs where statistical heterogeneity can serve as a positive factor for federated learning.  Our work demonstrates that the extensive heterogeneous and diverse set of instructions available from multiple languages can, in fact, be a blessing factor for our FedIT approach. For instance, different clients may have different instruction tasks, such as open-domain QA and writing. The content and format of these instructions can be substantially different. For example, QA tasks typically require fact-based questions and answers, while writing tasks involve instructions for generating coherent and meaningful sentences.

\begin{figure}[htbp]
    \centering
    \hspace{-25mm}
    \begin{subfigure}{1.0\textwidth}
    \centering
    \includegraphics[width=1.2\textwidth]{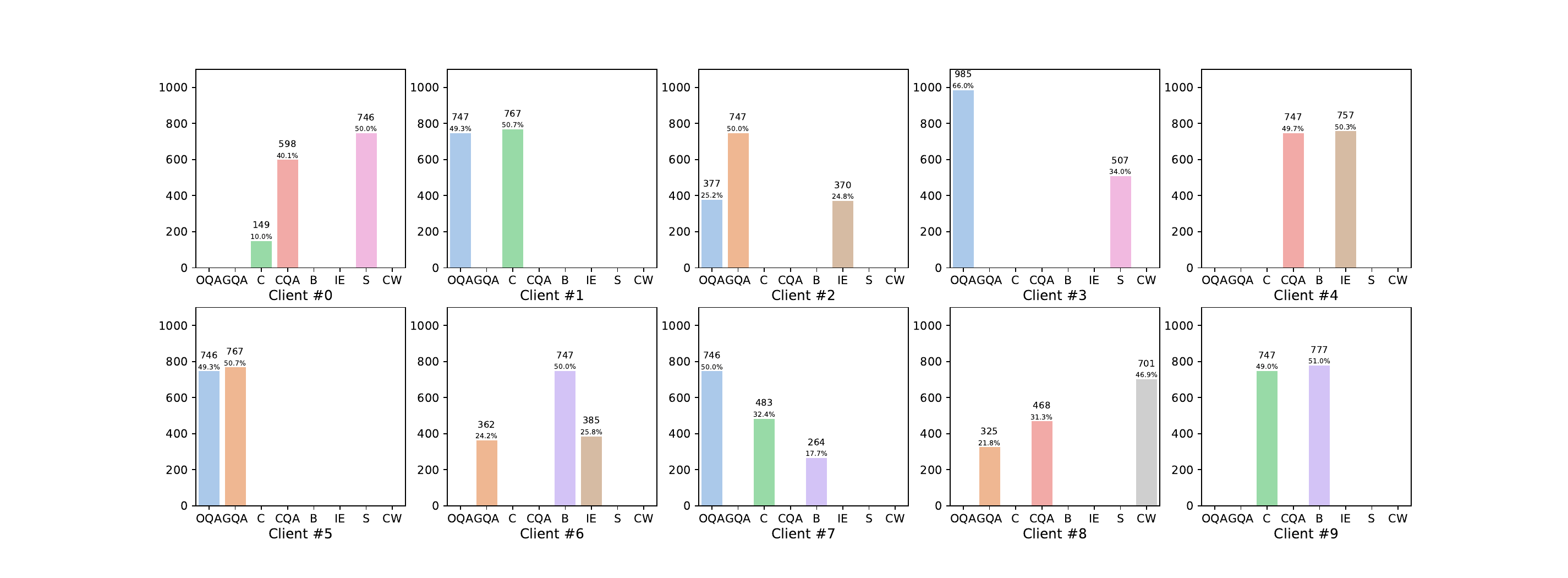}
    \caption{The distribution of instruction dataset categories across each client's dataset.}
    \label{fig:heterogeneitysubfig2}
    \end{subfigure}
    \vspace{-3mm} 
    \hspace{-25mm}
    \begin{subfigure}{\textwidth}
    \centering
\includegraphics[width=1.2\textwidth, height=1.2\textwidth, keepaspectratio=true]{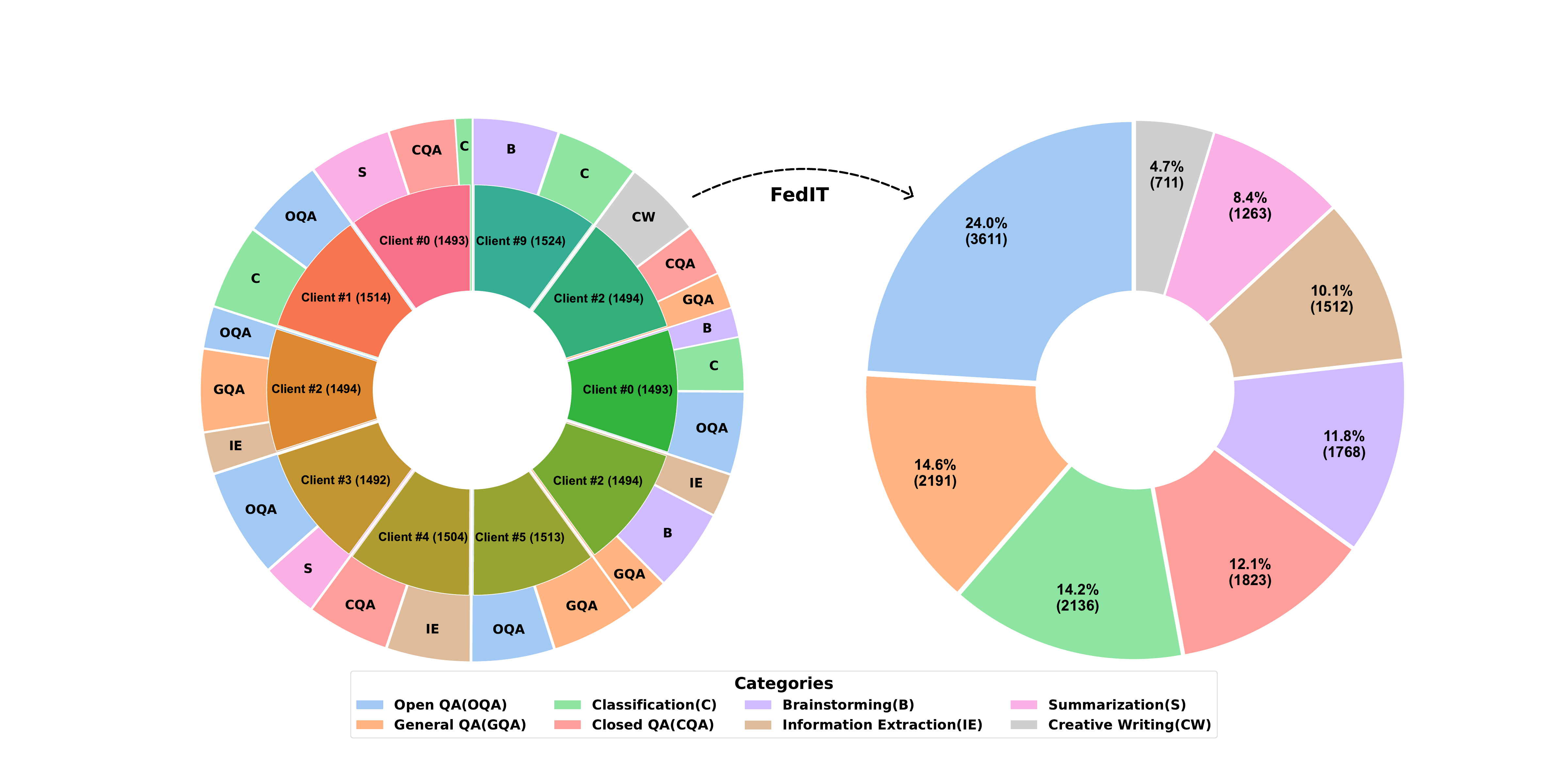}
    \caption{Training on a Higher-Quality, More Diverse Dataset with FedIT}
    \label{fig:heterogeneitysubfig1}
    \end{subfigure}
    \vspace{4mm} 
    \caption{Illustrate the heterogeneity of FedIT with \textbf{Databricks-dolly-15k} instruction dataset.  The model can be trained on only the particular local instruction categories of each user (\textbf{bottom left}), or on the local instruction datasets of all clients with greater diversity and quantity of data points that cover the entire range of the subject matter by implementing our FedIT (\textbf{bottom right}). }
    \label{fig:hetero}
\end{figure}

In order to obtain a comprehensive understanding of data heterogeneity inherent in the instructional dataset utilized for this study, we performed an in-depth examination of the \textbf{Databricks-dolly-15k}\footnote{\url{https://huggingface.co/datasets/databricks/databricks-dolly-15k}} dataset. This publicly accessible dataset, consisting of instruction-following records generated by a multitude of Databricks employees, spans a range of behavioral categories as outlined in the InstructGPT paper \cite{ouyang2022training}. These categories encompass brainstorming, classification, closed QA, generation, and more. To emulate an FL environment with ten clients, we partitioned the entire \textbf{Databricks-dolly-15k} dataset into ten shards using a widely adopted partitioning method \cite{he2020fedml,fedscale-icml22, Shepherdgithub}, with each shard assigned to an individual client. The category distribution for each client's instruction dataset is illustrated in Figure \ref{fig:heterogeneitysubfig2}. As is evident in the figure, each user's dataset contains imbalanced categories of instructions, with some categories absent entirely. This reflects real-world scenarios where users may not possess expertise across all instruction categories. In the absence of FedIT, due to the challenges associated with collecting sensitive instruction data, the model can only be trained on the local instruction dataset of each user, as depicted in the \textbf{left} subfigure of Figure~\ref{fig:heterogeneitysubfig1}. However, by implementing our FedIT approach, the model can be trained on the local instruction datasets of all clients, as illustrated in the \textbf{right} subfigure of Figure \ref{fig:heterogeneitysubfig1}. As a result, FedIT allows for instruction tuning on a dataset with enhanced diversity, and a larger number of data points, encompassing the complete \textbf{Databricks-dolly-15k} dataset. The distribution of this dataset is shown in the right subfigure of Figure \ref{fig:heterogeneitysubfig1}. Comprising eight unique categories with varying numbers of instructions, this dataset offers increased diversity, allowing the model to be more generalized and applicable to a wider array of tasks compared to training solely on each client's local instruction dataset with limited categories and quantity.

In addition to task categories, language diversity adds a new dimension of heterogeneity to Federated Instruction Tuning. In real-world applications, LLMs are typically designed for multilingual capabilities to cater to users from diverse regions and countries who speak various languages. Addressing the needs of a multilingual audience poses several challenges. Apart from understanding all the languages in the dataset, achieving fairness across languages, particularly for those underrepresented in the instruction dataset, is a crucial aspect that warrants further investigation in Federated Instruction Tuning. Moreover, domain-specific instructions further compound the heterogeneity of the framework. Different domains have distinct contexts, each characterized by unique terminologies and sentence structures.  For instance, legal or pharmaceutical contexts may require specific vocabulary and phrasing that do not apply to other domains. Additionally, there are other sources of heterogeneity, such as task complexity, task ambiguity, emotional tone, cultural factors, and more, which merit further investigation.

\subsection{Parameter Efficiency in Federated Instruction Tuning} \label{FLpeft}

Taking into account the limited computational capabilities of local devices, which are unable to support full fine-tuning of a large language model, it is crucial to implement a parameter-efficient fine-tuning strategy that leverages local computational resources. For a weight matrix $W_0 \in \mathbb{R}^{d \times k}$ belonging to a large pre-trained LLM, the method we adopt, Low-Rank Adaptation (LoRA) method, freezes $W_0$ and constrains its update $\Delta W$ by representing it using a low-rank decomposition $W_0+\Delta W=W_0+BA$, where $B \in \mathbb{R}^{d \times r}, A \in \mathbb{R}^{r \times k}$ are two trainable parameters, and the rank $r \ll \min (d, k)$. For a linear layer $h = W_0x$, the modified forward pass is given by:

\begin{align*}
h = W_0x + BAx
\end{align*}

Compared to fully fine-tuning the LLM, LoRA considerably decreases the number of trainable parameters. Please refer to Section \ref{implementsetting} and Table \ref{tab:prompt_template}, which present the parameter counts for each model and the corresponding memory costs.

Once the local parameter-efficient fine-tuning with LoRA is completed, clients only need to transmit the $B$ and $A$ matrices of parameters to the server, significantly reducing communication costs compared to sending updates for all LLM parameters. Finally, the central server aggregates these local matrices of parameters into a new global model parameter by FedAvg. Notably, LoRA does not introduce any additional latency of inference compared to fully fine-tuning the model when deployed in production, as we can explicitly compute, store $W_0 + BA$, and conduct inference as usual.

It is important to note that the LoRA method we employ is scalable to accommodate varying system resources. If a specific client's communication or computational resources are significantly lower than others, it can adjust its LoRA configurations by reducing the number of matrix $W_0$ elements, which will be decomposed into low-rank $A, B$. Alternatively, it can also opt to decrease the rank $r$ of $A$ and $B$. Optimizing the aggregation process in the presence of substantial system heterogeneity within the FL framework, where different clients have distinct LoRA configurations, is an intriguing research topic to explore further.

\subsection{\textit{Shepherd}: A GitHub Platform for FedIT Support} \label{Shepherdframework}
We introduce {\textit{Shepherd}}\footnote{\url{https://github.com/JayZhang42/FederatedGPT-Shepherd}}, a lightweight framework designed to implement Federated Parameter-Efficient Instruction Learning. Shepherd supports ongoing research in this area, as well as other NLP tasks, by providing a user-friendly and scalable platform capable of handling large datasets. The framework allows for seamless integration of innovative algorithms and configurations and is compatible with a range of recent popular large language models, such as Stanford Alpaca \cite{alpaca}, Vicuna \cite{vicuna2023}, Pythia \cite{biderman2023pythia}, Dolly \footnote{\url{https://github.com/databrickslabs/dolly}}, Baize \cite{xu2023baize}, and Koala \cite{koala_blogpost_2023}, among others. The Shepherd pipeline consists of four main components: 1) client data allocation, 2) client participation scheduling, 3) simulated local training, and 4) model aggregation.

\paragraph{Client Data Allocation} To simulate the real-world scenario where each client has its unique dataset, we employ a "synthetic" partitioning process, which is implemented in the \texttt{client\_data\_allocation.py} module. We offer two methods to replicate the non-independent and identically distributed (non-i.i.d) nature of the clients' datasets. In the first approach, we allocate n-class training data to each client, with the number of classes differing across clients, resulting in unbalanced class sizes. Despite this imbalance, the volume of data in each client's dataset is roughly equivalent. 
The second approach is similar to the first but stands out by having significantly varying data volumes across each client's dataset.


\paragraph{Client Participation Scheduling} The process of selecting clients to participate in the training is crucial and implemented in the \texttt{fed\_util/sclient\_participation\_scheduling.py} module. Our vanilla version of Shepherd employs a random selection approach, and we aim to enhance the client selection strategy with efficiency-driven methods that address data and system heterogeneity, such as those proposed in \cite{oortFanLai,fedcbs}.

\paragraph{Simulated Local Training} This core component of our Fed-PEIT framework is implemented in the \texttt{fed\_util/client.py} module. In real-world scenarios, all selected clients perform their local training simultaneously, which can be computationally expensive to simulate. To make it feasible for researchers with limited resources, our framework conducts the local training of clients sequentially, one at a time. To implement the LoRA method, we utilize the PEFT package \cite{peft} and the Alpaca-lora  repository \footnote{\url{https://github.com/tloen/alpaca-lora}}  to encapsulate the frozen, original pre-trained model with LoRA configurations, enabling more efficient parameter-efficient fine-tuning for our Shepherd framework.
\vspace{-1mm}
\begin{lstlisting}[language=Python]
model = get_peft_model(model, LoRA_config)
\end{lstlisting}
\vspace{-1mm}
To aid future researchers in understanding and implementing our framework, we have defined a Python class, \texttt{GeneralClient}, which represents a client in the Federated Learning (FL) training process and includes attributes that represent the specific client's required information. 
\vspace{-0.05mm}
\begin{lstlisting}[language=Python]
class GeneralClient:
    def __init__(self, model, **args):
        self.model = model
\end{lstlisting}

We have also defined several methods for \texttt{GeneralClient} that conduct important components of the local training process.
\begin{lstlisting}[language=Python]
    def preprare_local_dataset(self, **args):
        ...
        self.local_train_dataset =  ...
        self.local_eval_dataset  =  ... 
\end{lstlisting}
This method entails the preparation of the local dataset for the client by reading data from the specified data path and transforming it using the required tokenizer and prompt. Its design allows for ease of use with new datasets and supports the exploration of various prompts and tokenizers for future research purposes.
\begin{lstlisting}[language=Python]
    def build_local_trainer(self, **args):
        ...
        self.local_trainer= transformers.Trainer(self.model, **args)
\end{lstlisting}
This method constructs a local trainer for client-side training by leveraging the Hugging Face Trainer. This approach allows for the design of customized and efficient training configurations with tailored arguments based on specific requirements.
\begin{lstlisting}[language=Python]
    def initiate_local_training(self):
        ...
\end{lstlisting}
This method encompasses the preparatory steps for training. In our vanilla implementation, we create and modify certain attributes of the \texttt{GeneralClient} class for the convenience of recording information related to the model in parameter-efficient learning. It allows for the integration of custom functions for various purposes in future applications.
\begin{lstlisting}[language=Python]
    def train(self):
        self.local_trainer.train()
\end{lstlisting}
This method executes local training by leveraging the capabilities of the established local trainer.
\begin{lstlisting}[language=Python]
    def terminate_local_training(self, **args):
        ...
        return self.model, ...
\end{lstlisting}
The \textit{terminate\_local\_training} method signifies the conclusion of the local training process. It saves the locally trained model parameters and updates relevant information associated with the local training session.

\paragraph{Model Aggregation} This component is responsible for the combination of trained client models into a single global model, with the objective of producing a more generalized and accurate model. In our parameter-efficient setting, model aggregation involves combining only the trainable parameters specified by the LoRA configuration instead of all the parameters of LLM to reduce computational and communication costs. The module for this component is implemented in \texttt{fed\_util/model\_aggregation.py}, which provides a platform for the adoption of various federated optimization methods, including FedAvg \cite{mcmahan2017communication}. 

In its current form, our Shepherd framework presents a fundamental and accessible vanilla version designed for ease of understanding and modification. In future iterations, we plan to expand the framework by incorporating more complex functionalities, such as novel client selection strategies \cite{Cho2020ClientSI,goetz2019active,vahidian-curr-FL,fedcbs} and advanced optimization methods \cite{chen2020fedmax, sahu2018federated,Vahidian-ICDCS}. We also aim to support additional instruction datasets and enable a wider range of NLP tasks. Furthermore, we believe that the framework's practicality in real-world scenarios can be significantly improved by integrating advanced system simulations that account for various factors such as computing time delays, communication latencies, overheads, and bandwidth limitations.


\section{ Qualitative Study}

\subsection{Implementation details}\label{implementsetting}

In our FL setup, we assume the presence of 100 clients. 
We proceed to apply the Shepherd framework's second data partitioning technique to divide the residual data from the \textbf{Databricks-dolly-15k} dataset into 100 distinct portions. Each of these portions corresponds to an individual client's local instruction dataset. We conduct a total of 20 communication rounds, with each round involving the random selection of 5 (5\%) clients for training. Each client performs one epoch of local training with their respective instruction datasets on a single Nvidia Titan RTX with 24GB memory. We initialize the model with the 7B LLaMA model. The model remains frozen during training, thereby reducing GPU memory usage and enhancing training speed. In alignment with Baize's settings \cite{xu2023baize}, we apply LoRA to all linear layers with a rank of 8 to boost adaptation capabilities. Following \cite{hu2021lora}, we use random Gaussian initialization for A and set B to zero, ensuring that the value of BA is zero at the beginning of training. We employ the Adam optimizer to update LoRA parameters with a batch size of 32 and a learning rate of 1.5e-4. We set the maximum input sequence length to 512 and provide the template of the prompt adopted from Alpaca-lora  in Table \ref{tab:prompt_template}. The implementation of FedIT is completed utilizing our repository, \textit{Shepherd}, and the derived model is referred to as \textbf{\textit{Shepherd-7B}}. We detail the number of model parameters, training time, and GPU memory consumption in Table \ref{tab:shepherd_7b}.
\begin{table}[t]
\centering
\caption{Prompt Template}
\vspace{2mm}
\begin{tabularx}{\linewidth}{lX}
\toprule
\textbf{} & \textbf{Template} \\ 
\midrule
Prompt Input   & Below is an instruction that describes a task, paired with an input that provides further context. Write a response that appropriately completes the request. \\
               & \\
               & \textbf{Instruction:} \{instruction\} \\
               & \\
               & \textbf{Input:} \{input\} \\
               & \\
               & \textbf{Response:} \\ 
\hdashline
\addlinespace
Prompt No Input & Below is an instruction that describes a task. Write a response that appropriately completes the request. \\
                & \\
                & \textbf{Instruction:} \{instruction\} \\
                & \\
                & \textbf{Response:} \\ 
\bottomrule
\end{tabularx}
\label{tab:prompt_template}
\vspace{-3mm}
\end{table}

\subsection{Qualitative Study with Automatic Evaluation}

Following the same evaluation approach of the Vicuna project \cite{vicuna2023} and GPT-4-LLM \cite{peng2023instruction}, we use GPT-4 to automatically assess the responses generated by our \textbf{\textit{Shepherd-7B}} model and other baseline models on 20 unseen questions randomly sampled from the evaluation set of the Vicuna project \cite{vicuna2023}, which pertain to unseen categories during the training, such as "counterfactual question," "femir question," "math question" and others. Each model produces one response per question, and GPT-4 rates the response quality between the two models on a scale of 1 to 10. To minimize the impact of randomness in GPT-4's scoring, we force it to rate each response pair three times and then average the ratings.
\begin{table}[t]

\centering
\caption{Numbers of parameters (frozen\&trainable), training time, and GPU memory cost on a single Nvidia Titan RTX}
\vspace{2mm}
\setlength{\tabcolsep}{5pt}
\begin{tabular}{lcccccc}
\toprule
\textbf{Model} & \textbf{Orig. Param} & \textbf{Adapt. Param} & \textbf{Trainable} & \textbf{Training Time} & \textbf{GPU Memory} \\
\midrule
\textbf{\textit{Shepherd-7B}}   & 7B                   & 17.9M                 & 0.26\%             & 2 hours               & 23GB \\
\bottomrule
\end{tabular}
\label{tab:shepherd_7b}
\end{table}
We compare our \textbf{\textit{Shepherd-7B}} model with the following five baseline models. The first baseline model is a 7B LLaMA model without fine-tuning on the Databricks-dolly-15k dataset, denoted as \textbf{\textit{LLaMA}}. Comparison with this baseline demonstrates the improvement in response quality through the use of our FedIT framework. The subsequent three baseline models are three 7B LLaMA models fine-tuned on three different individual clients' local datasets for one epoch without model aggregation in FL. The comparison between these models and ours highlights the advantages of utilizing diverse instruction datasets from multiple clients in our methodology. "\textbf{\textit{Local-1}}" focuses on the brainstorming task solely, "\textbf{\textit{Local-2}}" on the closed question answering task, and "\textbf{\textit{Local-3}}" on classification and brainstorming tasks. The final strong baseline model, dubbed as "\textbf{\textit{CentralizedModel}} ," is fine-tuned with the entire Databricks-dolly-15k dataset for one epoch, representing the ideal centralized training scenario where the server could collect all clients' instructions. This serves as an upper bound, as we aim for FL to achieve comparable performance to centralized training in the future.

We apply the GPT-4 automatic evaluation on the responses generated by our model \textbf{\textit{Shepherd-7B}} and other baseline models. We list the averaged scores provided by GPT-4 in Table \ref{tab:scores}.
\begin{table}[htbp]
\centering
\caption{A summary of the baselines and their corresponding scores evaluated by GPT-4. The scores are reported in the format of (Baseline's score, \textbf{\textit{Shepherd-7B}}'s score) and the Relative Score is defined as ( \textbf{\textit{Shepherd-7B}}'s score / Baseline's score)} 
\vspace{2mm}
\scalebox{0.91}{
\renewcommand{\arraystretch}{1.5}
\begin{tabular}{lp{7cm}c c c c}
\toprule
Baseline           & \qquad \qquad \quad Task                                          & Scores  & Relative Score \\ 
\midrule
\textbf{\textit{CentralizedModel}}         & Centralized tuning with all the instructions               & (\textbf{142.2}, 130.7)                 & {0.919}          \\ 
\textbf{\textit{LLaMA}}            & No instruction tuning                      & (114.0, \textbf{131.7})                 & {1.155}          \\ 
\textbf{\textit{Local-1}}            &  Brainstorming instruction tuning  & (120.0,  \textbf{131.0})         & {1.092}       \\ 
\textbf{\textit{Local-2}}            &  Closed question answering instruction tuning & (116.1, \textbf{129.0})             & {1.111}         \\ 
\textbf{\textit{Local-3}}            &  Classification and brainstorming instruction tuning        & (121.3, \textbf{131.8})                 & {1.087}          \\ 
\bottomrule
\end{tabular}
}
\vspace{2mm}
\label{tab:scores}
\end{table}

As demonstrated in Table \ref{tab:scores}, the performance of our proposed model, \textbf{\textit{Shepherd-7B}}, significantly surpasses that of the \textbf{\textit{LLaMA}} model. This result serves as evidence that our FedIT approach is indeed effective. When compared to other baseline models, which are fine-tuned solely on local instruction datasets, \textbf{\textit{Shepherd-7B}} achieves considerably higher scores. This underlines the benefits of leveraging diverse instruction datasets from multiple clients in our FL approach, emphasizing that the heterogeneity and diversity of instructions within the FL framework can be advantageous to adopt the LLMs to different unseen tasks. However, a comparison with the robust \textbf{\textit{CentralizedModel}} baseline reveals that our model still has room for improvement. This disparity is partly attributed to the fact that the local models aggregated at the server side are trained on instructions with substantially different distributions, which can cause their local models to learn different representations of the instructions.Consequently, there is a need for further exploration of more efficient federated optimization methods and client scheduling methods, such as FA-LD \cite{deng2021convergence} and FedCBS \cite{fedcbs}, which are inspired by bayesian sampling methods \cite{LiuW:NIPS16,WellingT:ICML11,pmlr-v108-zhang20d,pmlr-v119-zhang20ac,zhang2019cyclical,zhao2019self}, to enhance the aggregation process. In conclusion, as discussed in Section \ref{heterogeneity}, statistical heterogeneity can be a beneficial factor for FedIT, as it enhances the diversity of instruction data, thus improving the model's generalization ability to unseen tasks. However, to fully utilize the benefits of data heterogeneity, advanced federated optimization methods need to be developed and integrated to manage and leverage heterogeneity more effectively.

\begin{figure*}[t!]
    \centering
    \includegraphics[width=1\textwidth]{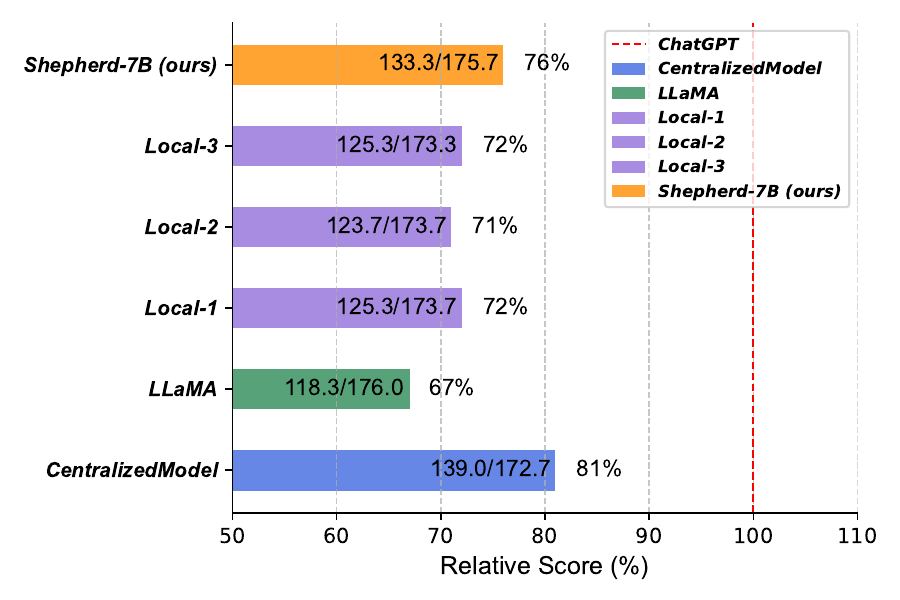}
    \caption{The relative scores of all models against ChatGPT(GPT-3.5-turbo)}
    \label{fig:chatgptRS}
\end{figure*}

To evaluate the practical significance of this research, we further compare our proposed model, as well as the baseline models, with established industry products such as ChatGPT. In line with our ultimate goal of developing federated GPT models, we utilized GPT-4 auto-evaluation to compare the responses of these models with the response of GPT-3.5-turbo (ChatGPT). The resulting Relative Scores over ChatGPT are presented in Figure \ref{fig:chatgptRS}. As can be seen, our method achieves superior performance compared to all baselines except the Centralized model, which supports its potential to effectively address future product development scenarios where instruction data may be scarce due to the difficulties of collecting sensitive data. Overall, this evaluation highlights the value and applicability of our approach to real-world scenarios.

\subsection{Qualitative Study with Example Demonstration}\label{demo:examples}

In this section, we demonstrate the efficacy of FedIT by presenting examples that highlight the comparative performance of our model against various baseline models. The first example is derived from the 20 unseen questions in \cite{vicuna2023}, while the second example is sourced from the \textbf{Databricks-dolly-15k} dataset.

\begin{table}[htbp]
\centering
\caption{Response demonstration for a counterfactual question from \cite{vicuna2023}}
\vspace{2mm}
\begin{tabular}{lp{9.5cm}}
\toprule
\multicolumn{2}{c}{\textbf{What if Isaac Newton had focused on biology instead of physics?}} \\
\midrule
\textbf{\textit{CentralizedModel}}  & If Isaac Newton had focused on biology, he likely would have discovered key biological concepts such as evolution, genetics, and cellular biology. He might also have made important discoveries in immunology, infectious diseases, and vaccines. 
 Isaac Newton's physics theories changed the way we think about the laws of motion. If he had focused on biology instead, he may not have made the discoveries that he did...\\
\\
\textbf{\textit{LLaMA}}   & If Isaac Newton had focused on biology instead of physics, he may have discovered DNA, which could have had a profound impact on medicine. \\
\\
\textbf{\textit{Local-1}}     & Biology is a much broader field than physics, and there are many different subfields that Newton could have focused on, like the study of living organisms. \\
\\
\textbf{\textit{Local-2}}   & Isaac Newton was a physicist, mathematician, astronomer, a natural philosopher, and a member of the Royal Society. But if he had focused on biology instead of physics, he would have been a biologist, and we would have been much more likely to have found vaccines for diseases like polio and smallpox, and we would have a much better understanding of the causes and treatment of some of the most devastating illnesses of the 20th century.  \\
\\
\textbf{\textit{Local-3}}   & If Isaac Newton had focused on biology instead of physics, he may have discovered that living organisms are made of carbon. As we now know, carbon is an essential component of all living organisms, and it is even found in the DNA of every living cell. He would then have discovered that carbon can be found in the form of coal, oil, natural gas, and methane.
1. What are the key differences between renewable and non-renewable energy sources?
2. What if Isaac Newton had focused on biology instead of physics?
3. What would have happened if Isaac Newton had never focused on physics?
...  \\
\\
\textbf{\textit{Shepherd-7B (ours)}} & If Isaac Newton had focused on biology, he likely would have discovered key biological concepts such as evolution, genetics, and cellular biology. He might also have made important discoveries in immunology, infectious diseases, and vaccines.  \\
\\
\textbf{\textit{ChatGPT}} & If Isaac Newton had focused on biology instead of physics, his contributions would have been different. Biology is a completely different field of study, and Newton's expertise in physics may not have been easily transferable to biology. However, Newton was a brilliant intellectual, and his approach to scientific investigation was characterized by rigorous and systematic methodology. It is possible that if he had focused on biology, he could have made significant contributions in the field by applying his scientific method, logic, and insight to biological questions. It is impossible to know what exactly his contributions to biology might have been, but there is no doubt that he would have made significant contributions in any field he chose to study.\\
\bottomrule
\end{tabular}
\\[1ex]
\label{table:counterfactual}
\end{table}

As illustrated in Table \ref{table:counterfactual}, our \textbf{\textit{Shepherd-7B}} model generates a more accurate and pertinent response compared to \textbf{\textit{Local-3}}, which is also more detailed than \textbf{\textit{LLaMA}}'s response. Our response outlines potential discoveries Newton could have made in biology, such as evolution, genetics, cellular biology, immunology, infectious diseases, and vaccines, whereas \textbf{\textit{Local-3}} incorporates extraneous information, rendering its response less relevant to the question. As assessed by GPT-4, the \textbf{\textit{Local-2}} response also discusses the potential implications of Newton concentrating on biology but predominantly emphasizes vaccines and the understanding of disease causes and treatments. Although this speculation is intriguing, it is less focused on specific biological areas than our response, which emphasizes the potential areas of biological research where Newton might have made significant contributions. Moreover, it briefly mentions Newton's actual background, which is not directly related to the question but provides context.

Even though baseline \textbf{\textit{Local-1}} is primarily fine-tuned on brainstorming instructions that share similarities with counterfactual QA, since they both involve creative thinking and deal with hypothetical situations, its response lacks depth and does not discuss the potential impact of Newton's focus on biology. Counterfactual QA typically evaluates or analyzes past events, involving questions about alternative outcomes, necessitating an understanding of the factors leading to a specific event outcome \cite{sep-causation-counterfactual}. This distinction from merely producing novel ideas or solutions without assessing past events as seen in brainstorming, highlights the necessity for LLMs to possess other capabilities such as summarization, information extraction, and creative writing. Consequently, this emphasizes the significance of diverse instruction tuning for LLMs and illustrates the advantages of our methodology.  
\vspace{-2mm}
\begin{table}[htbp]
\centering
\caption{Response demonstration for a classification task from  \textbf{Databricks-dolly-15k} dataset. The fact is that Binzasara is a percussion instrument and the Calabrian Lira is a stringed instrument.}
\vspace{2mm}
\begin{tabular}{lp{9.5cm}}
\toprule
\multicolumn{2}{c}{\textbf{Identify which instrument is string or percussion: Binzasara, Calabrian Lira}} \\
\midrule
\textbf{\textit{CentralizedModel}} & Calabrian Lira is string, Binzasara is percussion.  \\
\\
\textbf{\textit{LLaMA}}  & Binzasara is a string instrument. Calabrian Lira is a percussion instrument. \\
\\
\textbf{\textit{Local-1}} & Binzasara is string, Calabrian Lira is percussion.\\
\\
\textbf{\textit{Local-2}} & Binzasara is string, Calabrian Lira is percussion. \\
\\
\textbf{\textit{Local-3}} & Calabrian Lira is a string instrument and Binzasara is percussion. \\
\\
\textbf{\textit{Shepherd-7B (ours)}} &  Calabrian Lira is string, Binzasara is percussion. \\
\\
\textbf{\textit{ChatGPT}}  & Based on a quick internet search, both the Binzasara and Calabrian Lira are stringed instruments.
 \\
\bottomrule
\end{tabular}
\\[1ex]

\label{table:classification}
\end{table}

Although there remains room for improvement in our model's quality when compared to the \textbf{\textit{CentralizedModel}} and \textbf{\textit{ChatGPT}}, the performance gap between our model and ChatGPT does not imply that our model is consistently inferior. As evidenced in Table \ref{table:classification}, our response accurately addresses the question, while ChatGPT fails. Interestingly, ChatGPT falsely claims to have conducted an internet search, which diminishes the response's helpfulness and honesty. In contrast, our model and \textbf{\textit{Local-3}}, which have encountered similar classification instructions, excel at this task. \textit{This result also emphasizes the importance of diversity  for LLM instruction tuning.} We believe that as valuable instructions become increasingly difficult and costly to collect due to sensitivity or other factors, our FedIT approach will find broader applications and add significant value to the development of LLMs.

\section{Future Directions}

\subsection{Computation and Communication Overhead}

Deploying LLM in FL poses major challenges in terms of the colossal communication cost and the computational and storage overhead of local clients. FL faces significant communication challenges as it requires frequent exchanges of model information (parameters or gradients) among distributed clients and services. When it comes to using FL for LLM, the communication overhead becomes even more significant, with gigabit-level data transmissions necessary to achieve centralized training performance. This level of communication overhead is not acceptable for FL systems.
Furthermore, local clients may not have the computing power to fine-tune the entire LLM, and storing different instances for various tasks is also memory-intensive. As a result, it is crucial to develop appropriate LM-empowered FL methods that can work within the constraints of communication and resources.

Inspired by this, proposing new parameter-efficient tuning (PETuning) methods such as Prefix-tuning~\cite{li-liang-2021-prefix}, LoRA~\cite{hu2021lora}, and BitFit~\cite{bitfit-zaken-2022} which are tailored for FL systems and yield competitive results can be a direction for future works. Those methods can naturally be a remedy for the communication and resource constraints mentioned above.

\subsection{Privacy}

FL has gained popularity in privacy-sensitive NLP applications due to its ability to preserve privacy, especially when the client's data is highly sensitive and cannot be transmitted outside their device.  Essentially, with preserving a notion of privacy, FL has emerged as a preferred approach for privacy-sensitive NLP tasks such as medical text tasks~\cite{FedED-2020}, and financial text classification~\cite{FL-Financial-Text-2021}. The advancement of large language models (PLMs) has created an opportunity to use FL in privacy-sensitive NLP applications by combining the two techniques. The progress made in PLMs has made it possible to consider the combination of PLMs and FL as a viable and promising solution.

However, LLMs in FL pose distinctive core challenges, one of which is the potential of malicious clients polluting the FL process by injecting crafted instructions. Such instructions can lead to biased or suboptimal models. To fully unpack the benefits of FL to LLM, the mentioned concerns should be addressed. Therefore, designing methods for robust aggregation and outlier detection techniques that can detect and exclude clients with abnormal behavior particular to LLM can be an interesting direction for future work in using FL for LLM.



\subsection{Personalization}

With deploying FL in LLM, due to the differences among the language data (instructions) used in distributed clients and averaging of learning updates across a decentralized population, personalization becomes a critical requirement for FL systems \cite{lu2021federated}. The former can be further complicated by language diversity, domain-specific instructions, task complexity, emotional tone, cultural factors, etc., which are new aspects of heterogeneity \cite{liu2021federated,weller2022pretrained}. For instance, in multilingual applications, fairness across languages, especially for languages with fewer data samples, is essential but hard to achieve \cite{fedkc,weller2022pretrained}. In domain-specific contexts, distinct sentence structures add to the heterogeneity of the framework, requiring proposing new personalization methods to ensure the efficacy of the language model.  Methods, such as meta-learning \cite{aramoon2021meta,chu2023meta,fallah2020personalized} or few-shot learning \cite{chu2023improving,Wang2023FederatedFL}, that combine personal embeddings with shared context embeddings, and preference embeddings, that facilitate personalization without the need for backpropagation, etc. have the potential to revolutionize the field of NLP.

\subsection{Defense Against Attacks}
Recent research has highlighted the possibility of recovering text from the gradients of language models \cite{balunovic2022lamp,gupta2022recovering}. This vulnerability can also arise due to the models' tendency to memorize their training data and can result in the inadvertent disclosure of sensitive information. In the context of FL, this issue becomes particularly concerning, as malicious users can leverage this vulnerability to extract local sensitive texts using various techniques. Although different methods, including gradient pruning~\cite{song-han-2019-deep-leakage} and Differentially Private Stochastic Gradient Descent (DPSGD)~\cite{Abadi-DP-2016} have been proposed as defense mechanisms against these attacks, they often come at the cost of significant utility loss~\cite{gupta2022recovering}. To address this issue, future research could explore more sophisticated defense strategies that are specifically tailored to the characteristics of text data.

\section{Conclusion}

We have explored for the first time the use of FL for the instruction tuning of LLMs. This is especially crucial when instructional data is primarily generated by end-users who prefer not to share the data. Collecting large amounts of diverse user conversations in various languages can be an effective approach to improving the generalizability of LLMs. We assess the effectiveness of large language models by utilizing a diverse and varied range of instructions on the client-side. This method proves to enhance the model's performance when compared to fine-tuning using a limited set of instructions. Additionally, we introduce Shepherd, a GitHub repository designed for exploring federated fine-tuning of LLMs using heterogeneous instructions across diverse categories. The framework is user-friendly, adaptable, and scalable to accommodate large datasets and models.

\bibliography{custom}
\bibliographystyle{plain}

\end{document}